\documentclass[10pt,twocolumn,letterpaper]{article}

\usepackage{iccv}
\usepackage{times}
\usepackage{epsfig}
\usepackage{graphicx}
\usepackage{amsmath}
\usepackage{amssymb}
\usepackage{booktabs}
\usepackage{multirow}
\usepackage{enumerate}
\usepackage[numbers,sort]{natbib}

\usepackage{color}
\definecolor{MyDarkBlue}{rgb}{0.78,0.93,0.8}

\usepackage[pagebackref=true,breaklinks=true,letterpaper=true,colorlinks,bookmarks=false]{hyperref}

\iccvfinalcopy 

\def\iccvPaperID{0000} 

\ificcvfinal\fi
\begin{document}
	\bibliographystyle{plain}
	
	\title{Unsupervised Learning Framework of Interest Point Via Properties Optimization}
	
	\author{Pei Yan, Yihua Tan, Yuan Xiao, Yuan Tai, Cai Wen\\
		National Key Laboratory of Science and Technology on Multi-spectral Information Processing, \\School of Automation, 
		Huazhong University of Science and Technology, Wuhan 430074, China\\
		{\tt\small yanpei@hust.edu.cn yhtan@hust.edu.cn}
	}
	
	\maketitle

	\begin{abstract}
		This paper presents an entirely unsupervised interest point training framework by jointly learning detector and descriptor, which takes an image as input and outputs a probability and a description for every image point. The objective of the training framework is formulated as joint probability distribution of the properties of the extracted points. The essential properties are selected as sparsity, repeatability and discriminability which are formulated by the probabilities. To maximize the objective efficiently, latent variable is introduced to represent the probability of that a point satisfies the required properties. Therefore, original maximization can be optimized with Expectation Maximization algorithm (EM). Considering high computation cost of EM on large scale image set, we implement the optimization process with an efficient strategy as Mini-Batch approximation of EM (MBEM). In the experiments both detector and descriptor are instantiated with fully convolutional network which is named as Property Network (PN). The experiments demonstrate that PN outperforms state-of-the-art methods on a number of image matching benchmarks without need of retraining. PN also reveals that the proposed training framework has high flexibility to adapt to diverse types of scenes.
	\end{abstract}
	
	\section{Introduction}
	\label{section:introduction}
	
	Interest point is a sparse set of image point containing representative information \cite{balntas2017hpatches,mikolajczyk2005comparison,schmid2000evaluation}. With advantage of computation efficiency and representation stability, interest point has a wide utilization in matching tasks such as simultaneous localization and mapping, image registration and stereo reconstruction. Both computation efficiency and representation stability rely on the location and description of interest point which are produced by \emph{detector} and \emph{descriptor} respectively.
	
	All interest point related methods (including detectors and descriptors) can be classified as hand-crafted and learning based methods. Hand-crafted methods define some explicit criteria to extract the points \cite{harris1988combined,lowe2004distinctive,rosten2006machine} and calculate the descriptions while learning based methods are more flexible by training detector and descriptor according to some objectives \cite{strecha2009training,verdie2015tilde,detone2018superpoint}. Both hand-crafted methods and learning based methods make efforts to lead some important properties to interest point which benefit to diverse applications. 
	
	Currently, unsupervised learning is especially feasible to interest point by avoiding the tedious labeling of ground truth points \cite{savinov2017quad,ono2018lf}. Following the trend, this paper presents an entirely unsupervised training framework by jointly learning detector and descriptor, which takes an image as input and outputs a probability and a description for every image point. Here the probability reflects how likely the point become interest point, and the description is a feature vector used to represent the unique discriminability of an interest point. In essence, the learning model is to approach to the several optimal properties of interest point.  The joint probability distribution of the extracted point properties is used to formulate the objective of our training framework which is maximized to learn detector and descriptor through achieving all desired properties.
	
	The rest of this paper is organized as follows. In section \ref{section:related_work} we discuss the related work. In section \ref{section:framework}, we formulate our unsupervised learning framework of interest point. Section \ref{section:optimization} introduces latent variable to convert the objective function and explains the Expectation Maximization (EM) optimization algorithm. Section \ref{section:result} instantiate detector and descriptor with fully convolutional network, and show some experiments. Finally, in Section \ref{section:conclusion} we summarize this paper and list some possible future work.
	
	\section{Related Work}
	\label{section:related_work}
	The key idea of hand-crafted interest detector is to define explicit criteria to detect and describe interest point. Harris \cite{harris1988combined}, SIFT \cite{lowe2004distinctive}, SURF \cite{bay2006surf} and KAZE \cite{alcantarilla2012kaze} extract the points whose gray value have abrupt changes in two-dimensional space or scale space. FAST \cite{rosten2006machine} detects the points whose local gray distribution conforms to typical corner patterns. Gradient or gradient-like histograms are widely used in float-value description such as SIFT \cite{lowe2004distinctive}, SURF \cite{bay2006surf}, and Harris \cite{alcantarilla2012kaze}. Binary descriptors such as BREAF \cite{calonder2010brief},  BRISK \cite{leutenegger2011brisk}, FREAK \cite{Alahi2012FREAK} select some pixel pairs in neighborhood of interest point to calculate binary feature. Furthermore, scale and orientation estimations \cite{lowe2004distinctive,bay2006surf,alcantarilla2012kaze} are normally integrated into above descriptors to improve matching performance under scale and rotation transformations.
	
	TaSK \cite{strecha2009training} and TILDE \cite{verdie2015tilde} focus on the points extracted by existing detectors (i.e., Forstner \cite{verdie2015tilde} and SIFT \cite{lowe2004distinctive}) and train detectors to improve repeatability of these points under different illuminations. Quad-network \cite{savinov2017quad} approximates repeatability objective with ranking objective and doesn’t rely on any existing detectors. LIFT \cite{yi2016lift} first selects image patches around SIFT point \cite{lowe2004distinctive}, then train descriptor, orientation estimator and fine-tune detector. With known key point locations, SuperPoint \cite{detone2018superpoint} uses synthetic geometry shapes dataset to pre-trains a weak detector which is further trained in active learning way. Descriptor of SuperPoint is learned by another discriminability objective. In LF-Net \cite{ono2018lf} repeatability objective is approximated by calculating loss of detector between a pair of images, and its discriminability objective impact both descriptions and locations of extracted points, which is similar to LIFT \cite{yi2016lift}. SIPS \cite{cieslewski2018sips} fixes descriptor as existing model such as SIFT \cite{lowe2004distinctive}, and train detector to fit the descriptor. IMIP \cite{cieslewski2018matching} jointly learn detector and descriptor to extract the fixed number of points whose descriptions are limited as one-hot feature vectors.
	
	There are three essential differences between our training framework and existing methods. First, whereas existing methods improve some properties implicitly, we directly maximize the probability that interest point satisfies desired properties. Second, whether a point satisfies the required properties can be thought as latent binary variable so that our training framework can be optimized by Expectation Maximization algorithm. Third, our framework is very flexible to be generalized to any specific application by instantiating different models.
	
	\section{Unsupervised Learning Framework of Interest Point}
	\label{section:framework}
	\subsection{Problem Formulation}
	\label{subsection:framework_problem}
	Interest point is a sparse set of image point containing representative and discriminative information. In order to integrate both types of properties, a general framework learns \emph{detector} and \emph{descriptor} which extract interest point and its description respectively. First we introduce the general process of extracting interest point from a given scene.
	
	For the sake of describing the problem clearly, we give some notations. A actual world scene to acquire image is denoted as $I$. All possible viewpoint or illumination conditions for taking image are abstracted as transformation set $T=\{T_j \vert j\in Z\}$. Here $T_j$ represents a specific condition, and $Z=\{1,2,…,J\}$ represent all possible conditions. The image acquired from $I$ under condition $j$ is expressed as $T_j (I)$. Each point in $I$ and its corresponding mapped point in $T_j (I)$ are all denoted as $o_i$. Suppose the entire scene point set is $EP=\{o_i \vert i=1,2,…,N\}$. Here $N$ represents the number of scene points. 
	
	In general detector and descriptor take image $T_j (I)$ as input and output interest point and its description respectively. Detector $F$ is defined as a function outputting a probability $f_{ij}$ for every point $o_i$ in image $T_j (I)$,
	\begin{equation}
		f_{ij}=F(o_i,T_j(I)\big|\theta_F),
		\label{eq:detector}
	\end{equation}
	\noindent
	where $\theta_F$ are all the parameters of the model of detector, and $f_{ij}$ reflects how likely $o_i$ becomes an interest point. In practice probability threshold $Pt$ is introduced to obtain a deterministic interest point set. Interest point set of $T_j (I)$ is
	\begin{equation}
		IP_j=\{o_i\big|f_{ij}>Pt,o_i\in EP\}.
		\label{eq:interest_set}
	\end{equation}
	\noindent
	And $EP-IP_j$ is named as \emph{background point set}.
	
	Descriptor $D$ is defined as a function outputting a description vector $d_{ij}$ for every $o_i$ in image $T_j(I)$, i.e., 
	\begin{equation}
		d_{ij}=D(o_i,T_j(I)\big|\theta_D),
		\label{eq:descriptor}
	\end{equation}
	\noindent
	where $\theta_D$ are all the parameters of the model of descriptor, and $d_{ij}$ can be used to calculate the similarity between this point and other interest point, which is very important to determine the discriminability of an interest point. We always ensure $\left \| d_{ij} \right\|_2=1$ with length normalization. The description set of image $T_j (I)$ is denoted as
	\begin{equation}
		DS_j=\{d_{ij}\big|o_i\in IP_j)\}.
		\label{eq:description_set}
	\end{equation}
	
	The problem of learning based interest point is equivalent to learning the parameters of detector and descriptor. Whereas there is no unique supervised label for interest point and its description, it’s reliable to jointly train detector and descriptor in an unsupervised way which focuses on the essential properties of interest point. 
	Therefore, we propose an unsupervised training framework to optimize the learning model of interest point.
	
	The overview of the framework is shown as Figure \ref{fig:figure1}. With images acquired from the same scene, this framework jointly trains detector and descriptor. The training samples can be transformed from a scene image by simulating different illumination and viewpoint changes. Images are fed into detector and descriptor which outputs a probability and a description for every image point. Then, the joint probability of interest point properties is computed through the above two produced information. Finally, a proposed expectation maximization algorithm optimizes the joint probability to find the best model parameters, which means that the outputs of detector and descriptor have  achieved all desired properties.

	\begin{figure*}
		\begin{center}
			\includegraphics[width=1.0\textwidth]{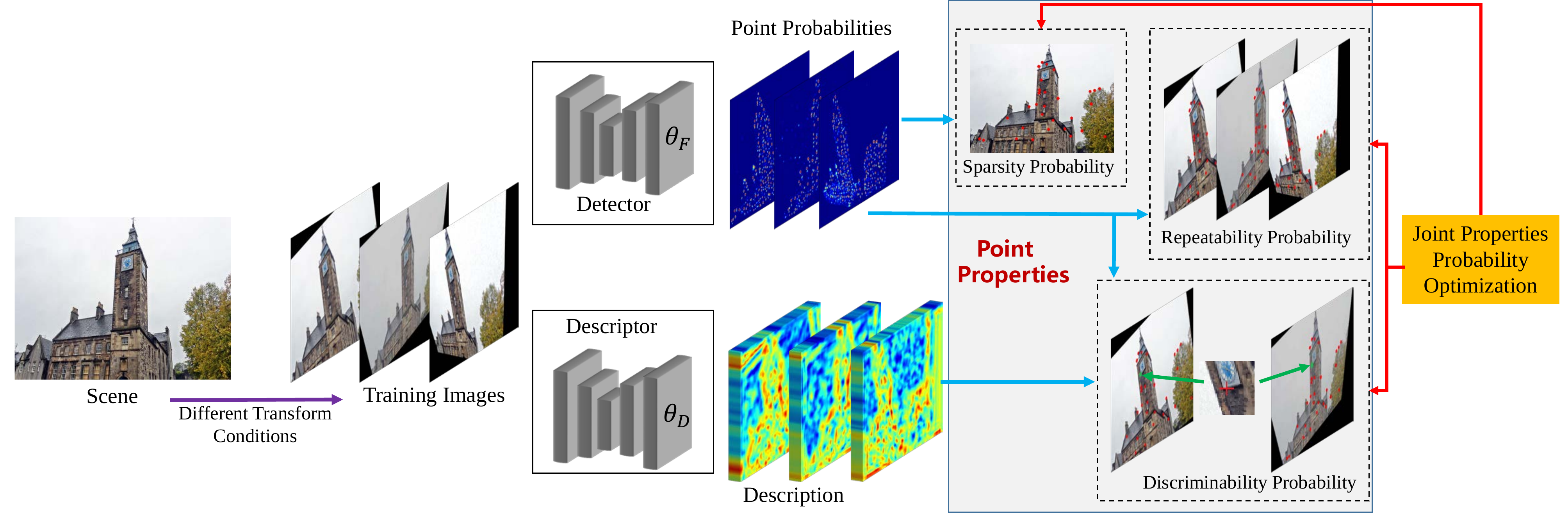}
		\end{center}
		\caption{Overview of unsupervised training framework via optimizing the properties of interest point. It consists of three parts: (1) training images transformed from a scene by simulating different imaging conditions, (2) models of detector and descriptor which produce interest point probability and description for each pixel, (3) Joint probability maximization algorithm that optimizes the properties of interest point.}
		\label{fig:figure1}
	\end{figure*}

	\subsection{Unsupervised Properties Optimization}
	\label{subsection:framework_overview}
	The probability that interest point of $T_j (I)$ satisfies given the $v$th property can be formulated as $P_v (IP_j,DS_j)$. Here $v\in \{1,2,…,V\}$ and $V$ is the number of all desired properties. Suppose all properties are independent and properties in different images are also independent. The objective maximizing the probability that interest point satisfies desired properties is
	\begin{equation}
		argmax_{\theta_F,\theta_D}\prod_{j}\prod_{v}P_v(IP_j,DS_j).
		\label{eq:general_obj}
	\end{equation}
	In this paper we make $V=3$, and \emph{sparsity}, \emph{repeatability} and \emph{discriminability} are selected as essential properties in this paper. The probabilities that an interest point satisfies above properties are denote as \emph{sparsity probability}, \emph{repeatability probability} and \emph{discriminability probability}, whose formulations are introduced in subsequent subsections.
	
	\subsection{Sparsity Probability}
	\label{subsection:framework_sparsity}
	Sparsity is an essential property controlling the limited number of interest points sparsely scatter over image. According to definition sparsity only rely on interest point rather than its description, so its probability can be represented as $s(IP_j)$.
	
	As a specific implementation, we define $s_{loc} (IP_j )$ as the probability that interest point set of image $T_j (I)$ is locally sparse. Local sparsity indicates that there is none in the interest point $o_i$’s neighborhood except itself. $U(o_i )$ represents the neighborhood of point $o_i$ whose radius is $rad$. Here $rad$ is a small integral (e.g., $rad=4$), and $U(o_i)$ doesn’t contain $o_i$ itself. Define $\left \| \cdot \right \|$ returns the number of elements in a given set. Then the local sparsity probability of $o_i$ is defined as
	\begin{equation}
		s_{loc}(o_i)=
		\begin{cases} 
			1,  & \left \| \{ o_{i'}\big|o_{i'}\in U(o_i),o_{i'}\in IP_j \} \right \|=0 \\
			0, & otherwise
		\end{cases}.
		\label{eq:local_sparsity}
	\end{equation}
	Suppose the local sparsity probabilities of different interest points is independent, the local sparsity probability of entire interest point $IP_j$ is
	\begin{equation}
		s_{loc}(IP_j)=\prod_{i\in IP_j}s_{loc}(o_i).
		\label{eq:joint_local_sparsity}
	\end{equation}
	Except making interest point to be sparse locally, sparsity is also a global properties controlling the number of interest points.  Depending on the number of interest points $\left \| IP_j \right \|$, $s_{num} (\left \| IP_j \right \|)$ represents the probability that number of interest point is reasonable.  We define it as
	\begin{equation}
		s_{num}(\left \| IP_j \right \|)=
		\begin{cases} 
			1,  & N_{min}<\left \| IP_j \right \|<N_{max} \\
			0, & otherwise
		\end{cases}.
		\label{eq:num_sparsity}
	\end{equation}
	Combining Equation \ref{eq:joint_local_sparsity} and \ref{eq:num_sparsity},  the sparsity probability is expressed as
	\begin{equation}
		s(IP_j)=s_{loc}(IP_j)\cdot s_{num}(\left \| IP_j \right \|).
		\label{eq:sparsity}
	\end{equation}
	\subsection{Repeatability Probability}
	\label{subsection:framework_repeatability}
	Repeatability indicates how likely a point can be extracted repeatedly under different viewpoints and illuminations. With this definition repeatability is determined on multiple images acquired from the same scene rather than a single image, so repeatability is defined on $I$ to simplify notations. Denote repeatability probability of point $o_i$ in $I$ as $r_i$, which represents the probability that $o_i$ is extracted in $I$, i.e.
	\begin{equation}
		r_i=\lim_{J \to \infty}\left( \frac{1}{J}\sum_j f_{ij}\right).
		\label{eq:point_repeat}
	\end{equation}
	where $f_{ij}$ is the probability output by detector for $o_i$ in image $T_j (I)$ which is defined in Equation \ref{eq:detector}. Then $1-r_i$ is the probability point $o_i$ belongs to background point set. In the reminder of the paper we ignore the limit of $J$ to simplify notations. Suppose the repeatability of each point is independent on each other, the repeatability probability of $IP_j$ satisfies
	\begin{equation}
		r(IP_j)= \prod_{o_i\in IP_j}r_i\prod_{o_i\in EP-IP_j}(1-r_i).
		\label{eq:repeat}
	\end{equation}
	\subsection{Discriminability Probability}
	\label{subsection:framework_discriminability}
	Discriminability denotes how likely an interest point in one image is more similar to the same point than the other interest points in another image. The similarity between point $o_i$ and $o_{i'}$ is normally defined as the inner product of their description vector $d_{ij}$ and $d_{i'j'}$, which is formulated as
	\begin{equation}
		sim_{iji'j'}=\left \langle d_{ij}, d_{i'j'}\right \rangle.
		\label{eq:similarity}
	\end{equation}
	$sim_{iji'j'}\in [-1,1]$ because $\left \| d_{ij} \right \| _2=1$. 
	If $i=i'$, $sim_{iji'j'}$ denotes the similarity between the same point in two images, which is termed as \emph{positive pair}. Otherwise  when $i \not =i'$, $sim_{iji'j'}$  represents the similarity between different points in two images, which is termed as \emph{negative pair}. 
	
	Denote indicator function as $\mathbb I (\cdot)$ which return 1 if and only if logical operation is true. Define function $max$ returning the maximum of a set and returning $–Inf$ for empty set. Then the discriminability probability of an interest point is
	\begin{equation}
		\begin{aligned}
			c_i=
			& \frac{1}{J(J-1)}\sum_j \sum_{j' \not= j}\mathbb I \left(sim_{ijij'}> \right. \\
			& \left. max(\{sim_{iji'j'} \big|o_{i'}\in IP_{j'},i'\not=i \})) \right) \\
		\end{aligned},
		\label{eq:discri_ori_point}
	\end{equation}
	\noindent
	where $o_{i'}$ must be an interest point rather than an arbitrary point because only interest point will be considered in matching process. In practice Equation \ref{eq:discri_ori_point} is too sharp to represent the gap between current and optimal discriminability. So we approximate it with
	\begin{equation}
		\hat{c_i}=exp(\alpha (h_i-H)),
		\label{eq:discri_point}
	\end{equation}
	where $h_i$ is the difference between similarity of positive pair and negative pairs, and $H$ is the maximum of $h_i$. $\alpha$ is a factor controlling the sensitivity of discriminability probability $\hat{c}_i$ with respect to $h_i$. The formulation of $h_i$ is
	\begin{equation}
		\begin{aligned}
			h_i=
			& \frac{1}{J(J-1)}\sum_j \sum_{j' \not= j}\bigl( min(m_p,sim_{ijij'})- \\
			& \frac{\lambda}{\left \| IP_{j'}-o_i \right \|}\sum_{o_{i'}\in IP_{j'}-o_i}max(m_n,sim_{iji'j'}) \bigr)
		\end{aligned}.
		\label{eq:discri_diff}
	\end{equation}
	The formulation of $h_i$ is inspired by descriptor loss in [7]. Here $m_p \in [-1,1]$ and $m_n \in [-1,1]$ are named as positive margin and negative margin, which can be seemed as the target similarity of positive pair and negative pair. $\lambda$ is a weight to balance positive pair and negative pair. Because $sim_{iji'j'} \in [-1,1]$ the maximum of $h_i$ is $H=m_p-\lambda m_n$.
	
	Suppose the discriminability of each point is independent, the discriminability  probability of $IP_j$ is
	\begin{equation}
		\hat{c}(IP_j,DS_j)=\prod_{o_i \in IP_j}\hat{c}_i.
		\label{eq:discri}
	\end{equation}
	\noindent
	Note in Equation \ref{eq:discri} we don’t concern discriminability between interest point and background point, or discriminability between background points themselves.
	
	\subsection{Objective of Properties Optimization}
	\label{subsection:framework_objective}
	During training, we generally have a scene set $\{I^k\vert k=1,2,...,K\}$. Denote interest point set of image $T_j (I^k )$ as $IP_j^k$ where the descriptor of each point $o_i$ is  $d_{ij}^k$. Thus, description set of image $T_j (I^k)$ is $DS_{kj}=(d_{ij}^k\vert o_i \in IP_{kj})$. Notations $r_i$, $h_i$, $\hat{c}_i$ are also extended as $r_i^k$, $h_i^k$, $\hat{c}_i^k$ to represent repeatability and discriminability for point $o_i$ in scene $I^k$. Then the objective of properties optimization is
	\begin{equation}
		argmax_{\theta_F,\theta_D}\prod_k\prod_j s(IP_j^k)\cdot r(IP_j^k)\cdot \hat{c}(IP_j^k,DS_j^k),
		\label{eq:ori_obj}
	\end{equation}
	which is named as \emph{properties objective}. Note the description set $DS_j^k$ is ignored in $s$ and $r$ because sparsity and repeatability probability don’t concern description of interest point.

	\section{Optimization and Implementation}
	\label{section:optimization}
	\subsection{Problem Conversion with Latent Variable}
	\label{subsection:optimization_conversion}
	
	It’s hard to straightforward optimize properties objective with conventional gradient based algorithm. In Equation \ref{eq:interest_set} interest point set $IP_j$ is determined by the probability threshold $Pt$ where the derivative of the logical operation doesn’t exist. Therefore, we introduce latent variable to solve this problem.
	
	Binary latent variable $y_i^k$ is formulated for every point $o_i^k$ in scene $I^k$ and $y_i^k=1$ if and only if point $o_i$ in any image $T_j (I^k)$ satisfy all desired properties. $y^k$ is the vector whose $i$th component is $y_i^k$, and $Y$ is the matrix whose item in $k$th row and $i$th col is $y_i^k$. Point $o_i^k$ is defined as \emph{satisfied point} if $y_i^k=1$, and satisfied point set of $I^k$ is $\{o_i\vert y_i^k=1\}$.
	
	In original objective \ref{eq:ori_obj} $\theta_F$ and $\theta_D$ are optimized to make interest point sets $IP_j^k$ and their descriptions achieve the desired properties, meaning that optimal solution of $IP_j^k$ must be the point set $\{o_i^k\vert y_i^k=1\}$. So replace $IP_j^k$ as $\{o_i^k\vert y_i^k=1\}$ in Equation  \ref{eq:ori_obj} won’t change its optimal solution. We first convert the property probability to more compact formulations with $y_i^k$.
	The number of satisfied points can be calculated with $\left \| y^k \right \| =\sum_i y_i^k$ . Redefine local sparsity probability for $y_i^k$ as
	\begin{equation}
	s_{loc}(y_i^k)=
	\begin{cases} 
	y_i^k,  & \left( \sum_{o_{i'} \in U(o_i)}y_{i'}^k \right)=0 \\
	0, & otherwise
	\end{cases}.
	\label{eq:local_sparsity}
	\end{equation}
	For vector $y^k$ we define the result of $s_{loc} (y^k ) $ is still a vector, whose $i$th component is $s_{loc} (y_i^k )$. 
	
	Redefine sparsity probability of $y^k$ as
	\begin{equation}
		s\left(y^{k}\right)=\mathbb{I}\left(N_{\min }<\left\|y^{k}\right\|<N_{\max }\right) \cdot \mathbb{I}\left(s_{l o c}\left(y^{k}\right)=y^{k}\right)
		\label{eq:spa_redef}
	\end{equation}
	The conversion of repeatability probability is straightforward.
	\begin{equation}
		r\left( y^k \right) =\prod_i{\left(r_{i}^{k}\right)}^{y_{i}^{k}}\left( 1-r_{i}^{k} \right) ^{1-y_{i}^{k}}
		\label{eq:rep_redef}
	\end{equation}
	Also we can redefine discriminability probability as
	\begin{equation}
		\begin{aligned}
		h_i^k(y^k)=
		&\frac{1}{J(J-1)}\sum_{j}\sum_{j' \ne j} \bigl( min\left( m_p,sim_{iji'j'} \right)-\\
		&\frac{\lambda}{\left\| y^k \right\|}\sum_{i' \ne i}{y_{i'}^k}max\left(m_n,sim_{iji'j'}\right) \bigr)
		\end{aligned}
		\label{eq:diff_redef}
	\end{equation}
	\begin{equation}
		\hat{c}\left(y^{k}, D S_j^{k}\right)=\prod_{i} \exp \left(\alpha y_i^k\left(h_i^k\left(y^{k}\right)-H\right)\right)
		\label{eq:disc_redef}
	\end{equation}
	Note $\hat{c}(y^{k}, D S_j^{k})$ for any $j$ is same because all images of $I^k$ share the same $y^k$, so we replace it with $\hat{c}(y^k)$ in this case. In fact, all property probabilities for different images of $I^k$ are exactly equal by sharing $y^k$. Then properties objective \ref{eq:ori_obj} can be converted as \emph{latent properties objective}
	\begin{equation}
		argmax_{\theta_{F}, \theta_{D}, y} \prod_{k} s\left(y^{k}\right) \cdot r\left(y^{k}\right) \cdot \hat{c}\left(y^{k}\right).
		\label{eq:obj_redef}
	\end{equation}
	\noindent
	According to Equation \ref{eq:spa_redef} sparsity probability $s(y^k)$ is either 1 or 0. To maximize latent properties objective $s(y^k)$ has to be ensured as 1, so that objective \ref{eq:obj_redef} is equivalent to
	\begin{equation}
		\left\{\begin{array}{c}{\operatorname{argmax}_{\theta_{F}, \theta_{D}, y} \prod_{k} r\left(y^{k}\right) \cdot \hat{c}\left(y^{k}\right)} \\ {s . t . N_{\min }<\left\|y^{k}\right\|<N_{\max }, s_{l o c}\left(y^{k}\right)=y^{k}}\end{array}\right.
		\label{eq:latent_obj}
	\end{equation}
	\noindent
	With logarithm transformation it’s equivalent to
	\begin{equation}
		\left\{\begin{array}{c}{\operatorname{argmax}_{\theta_{F}, \theta_{D}, y} L=\sum_{k} \sum_{i} l_i^k} \\ {s . t . N_{\min }<\left\|y^{k}\right\|<N_{\max }, s_{l o c}\left(y^{k}\right)=y^{k}}\end{array}\right.
		\label{eq:log_latent_obj}
	\end{equation}
	\noindent
	where log-likelihood item
	\begin{equation}
		l_i^k=y_i^k \log r_i^k+\left(1-y_i^k\right) \log \left(1-r_i^k\right)+\alpha y_i^k\left(h_i^k\left(y^{k}\right)-H\right)
		\label{eq:log_item}
	\end{equation}
	\noindent
	In objective \ref{eq:log_latent_obj} $L$ is named as log-likelihood function.
	
	\begin{figure*}
		\begin{center}
			\includegraphics[width=1.0\textwidth]{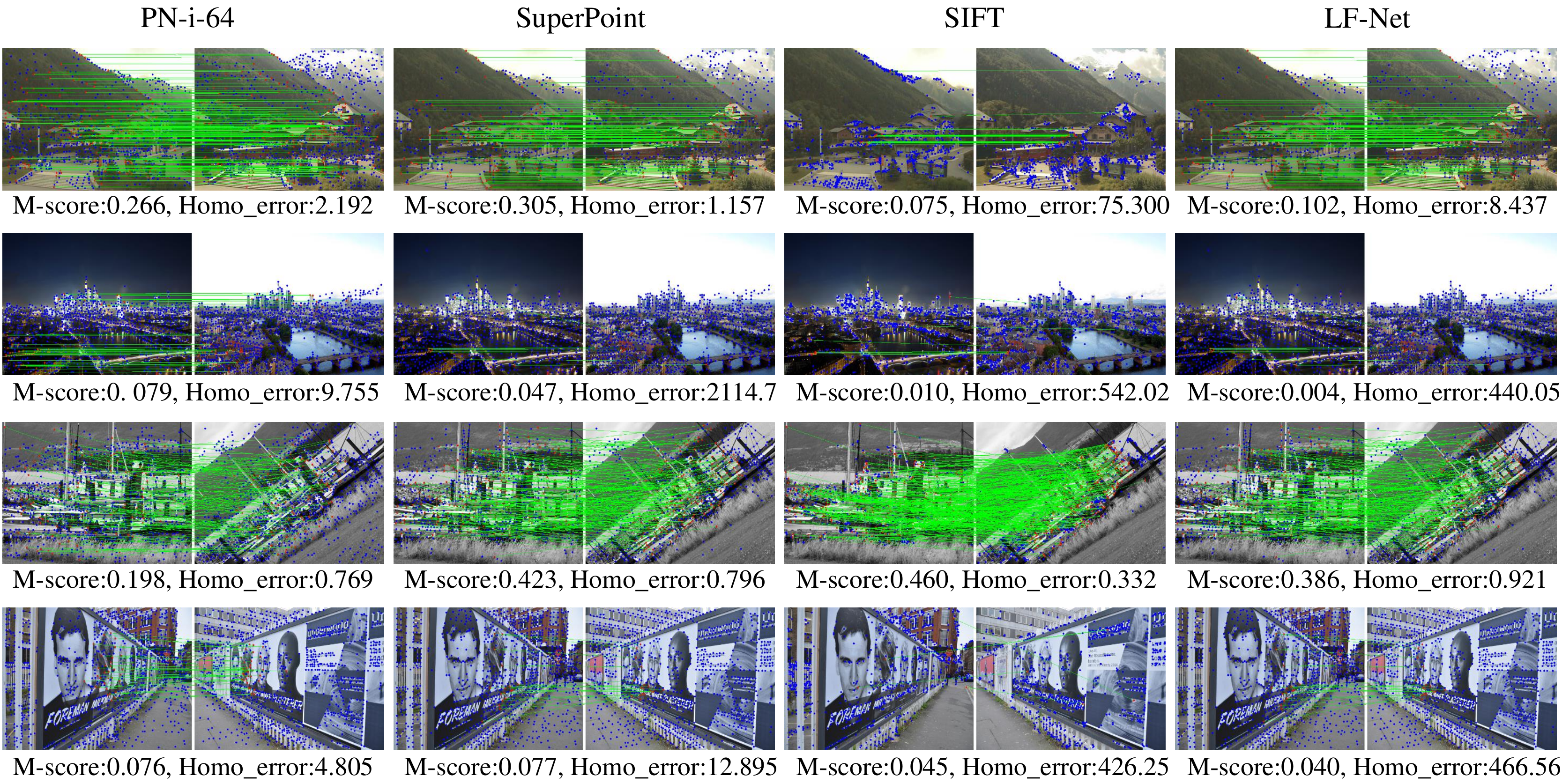}
		\end{center}
		\caption{Visual matching results of state-of-the-art algorithms and our PN-i-64 model (M-score indicates Matching Score, and Homo-error means the error of estimated Homography) . First col: our PN-i-64 model, second col: SuperPoint, third col: SIFT and fourth col: LF-Net. All points extracted by each method are shown in blue dots whereas correct matches are shown in green lines and red dots. M-score indicates Matching Score that higher is better, and Homo-error means the error of estimated Homography that lower is better. Whereas SuperPoint and SIFT achieve best performance in some specific scenes, PN-i-64 model give reliable results under both illumination and viewpoint changes.}
		\label{fig:figure2}
	\end{figure*}
	
	\subsection{Expectation Maximization of latent properties objective}
	\label{subsection:optimization_EM}
	Objective \ref{eq:log_latent_obj} can be optimized with Expectation Maximization algorithm (EM). Suppose the optimization contains $T$ times iterations. Let $\theta_F^0$ and $\theta_D^0$ are initial parameters, and the parameters of detector and descriptor are $\theta_F^t$ and $\theta_D^t$ after $t$ times iterations. 
	
	\noindent
	\emph{E-step:}
	
	In E-step of iteration $t$, we need to obtain the expectation of log-likelihood function $L$ with respect to $y$, which is denoted as $E_y(L)$. To achieve it we first estimate the probability distribution $P(y\vert \theta_F^{t-1},\theta_D^{t-1})$. 
	Because different $y^k$ are independent we only discuss the distribution of $y^k$ directly. The formulation for probability of $y^k$ is
	\begin{equation}
		\left\{\begin{array}{c}{P\left(y^{k} | \theta_{F}^{t-1}, \theta_{D}^{t-1}\right)=r\left(y^{k}\right) \cdot \hat{c}\left(y^{k}\right)} \\ {s . t . N_{\min }<\left\|y^{k}\right\|<N_{\max }, s_{l o c}\left(y^{k}\right)=y^{k}}\end{array}\right..
		\label{eq:prob_y}
	\end{equation}
	Denote all possible $y^k$ form the set
	\begin{equation}
		Y^k=\{ y^k \big| N_{min}<\left\| y^k \right\|<N_{max},s_{loc} (y^k)=y^k \}
	\end{equation}
	Then the probability distribution of $y_i^k$ is
	\begin{equation}
		P\left(y_i^k=1 | \theta_{F}^{t-1}, \theta_{D}^{t-1}\right)=\frac{1}{Z}\sum_{y^k \in Y^k}  y_i^k \cdot r\left(y^{k}\right) \cdot \hat{c}\left(y^{k}\right),
		\label{eq:solu_y}
	\end{equation}
	where $Z=\sum_{y^k \in Y^k} r(y^{k}) \cdot \hat{c}(y^{k})$ is the normalization factor. Denote $p_i^k=P(y_i^k=1)$, then the expectation of $y_i^k$ is $p_i^k$. So the expectation of $L$ is
	\begin{equation}
		E_{y}(L)=\sum_{k} \sum_{i} E_{y_{i}^{k}}\left(l_i^k\right)
		\label{eq:expect_L}
	\end{equation}
	Here 
	\begin{equation}
		\begin{aligned}
			E_{y_i^k}\left(l_i^k\right)=
			& p_i^k \log r_i^k+\left(1-p_i^k\right) \log \left(1-r_i^k\right)+ \\
			& \alpha E_{y^k}(y_i^k \cdot h_i^k(y^k)) -\alpha p_i^k H
			\label{eq:expect_item}
		\end{aligned}
	\end{equation}
	
	\noindent
	\emph{M-step:}
	
	In M-step $\theta_F^t$ and $\theta_D^t$ are obtained by maximizing the expectation of $L$. Normally the number of parameters is very huge, so Gradient Ascent algorithm (GA) is selected to achieve M-step in this paper. The gradient is computed as:
	\begin{equation}
		\frac{\partial L}{\partial \theta^{t-1}_F} = \sum_k \sum_i \frac{p^k_i-r^k_i}{J \cdot r^k_i \cdot (1-r^k_i)} \sum_j \frac{\partial F(i,T_j(I^k))}{\partial \theta ^{t-1}_F} 
		\label{eq:grad_F_1}
	\end{equation}
	\begin{equation}
		\frac{\partial L}{\partial \theta^{t-1}_D} = \alpha \sum_k \sum_i  \frac{\partial E_{y^k}(y_i^k \cdot h_i^k(y^k))}{\partial \theta_D^{t-1}}
		\label{eq:grad_D_1}
	\end{equation}
	
	Similarity $sim_{iji'j'}$ is computed with inner product which is derivable. So if the detector $F$ and descriptor $D$ are derivable, their parameters $\theta_F$ and $\theta_D$ can be optimized with GA. Theoretically, from $\theta_F^t$ and $\theta_D^t$ to $\theta_F^{t+1}$ and $\theta_D^{t+1}$ GA may need multiple times updates to achieve convergence. In practice, both Equation \ref{eq:solu_y} and \ref{eq:grad_D_1} lead to very high computational complexity, so we introduce Mini-Batch Approximation as an efficient implementation. All details of of this approximation are outlined in Supplementary Section 1.
	
	\begin{figure*}
		\begin{center}
			\includegraphics[width=1.0\textwidth]{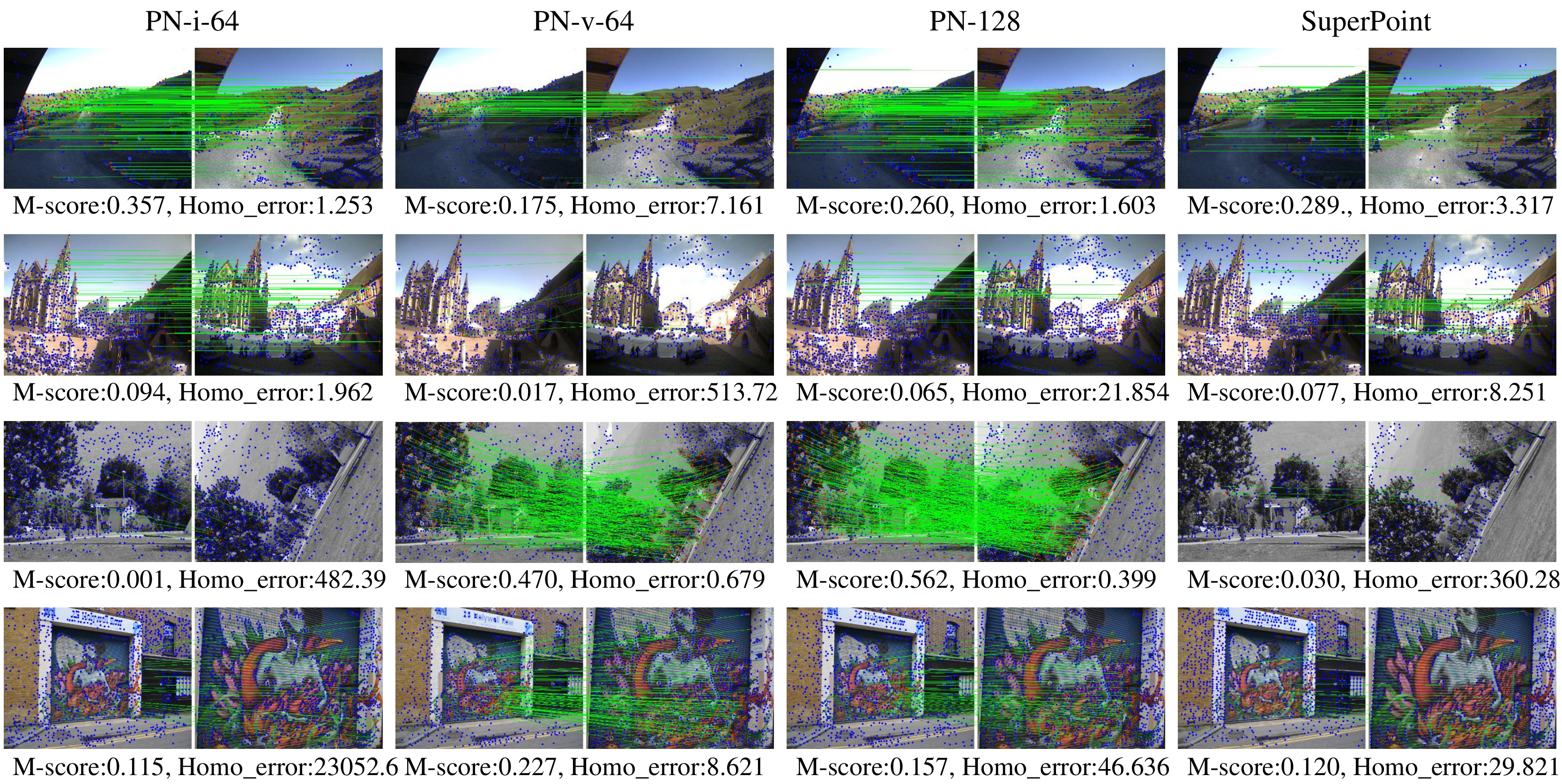}
		\end{center}
		\caption{Visual matching results of Superpoint and different Property Networks. First col: PN-i-64 model, second col: PN-v-64 model, third col: PN-128 model and fourth col: SuperPoint. Here all notations are same to what in Figure 2. By focusing on sharp illumination changes, PN-i-64 has significant superiority for scenes with illumination changes. Analogously PN-v-64 achieve most reliable performance under large range of viewpoint changes. PN-128 can manage diverse changes by learning with both sharp illumination changes and large range of viewpoint changes. With description vector length of 128, PN-128 achieves similar performance to state-of-the-art SuperPoint whose description vector length is 256.}
		\label{fig:figure3}
	\end{figure*}
	
	\section{Experimental Result}
	\label{section:result}
	\subsection{Experiment Setup}
	\label{subsection:result_setup}
	Our framework is flexible to be integrated with different models. In this paper we implement detector $F$ and descriptor $D$ as two Fully Convolutional Network \cite{long2015fully,zeiler2010deconvolutional} with Batch Normalization \cite{ioffe2015batch}, and parameters of their encoders are shared. We name this implementation as Property Network (PN). The architecture of PN is demonstrated in  Supplementary Section 2.
	
	In this paper we select MS-COCO 2014 \cite{Lin2014Microsoft} as training dataset, which comprises of more than 80 thousand images. We treat each single image as a scene which is transformed to the training images with different illuminations and viewpoints through simulated transformations.  The details of simulated transformations are outlined in Supplementary Section  3. 
	
	To adapt to different kinds of illumination and viewpoints changes, we train three different PN models corresponding to three transform simulation conditions. PN-i-64 model is trained with the condition of sharp illumination change and medium viewpoint change. PN-v-64 is trained with the condition of large range viewpoint change and medium illumination change. PN-128 is trained with the condition of sharp illumination change and large range viewpoint change. The architectures of above three networks are identical except the length of their description vector. The length of description vector of PN-i-64, PN-v-64 and PN-128 are 64, 64, 128 respectively.
	
	The configurations of all hyperparameters of PN are as below. During training we resize all images to $320\times240$ and stack different images to mini-batches (but the size of testing image can be arbitrary). In every iteration two scenes are randomly transformed ten times respectively, so there are in fact twenty images in a mini-batch (i.e., $B=2$ and $J=10$). In entire training we fix number range $N_{min}=200$, $N_{max}=400$, and non-maximum suppression radius $rad=4$ pixels. For discriminability, negative pair weight $\lambda=10/N_{max}$ , positive margin $m_p=1$, negative margin $m_n=0.2$ and discriminability weight $\alpha=1$. We always stop training after two epochs. The FCN model is implemented with and solved by Adam optimizer \cite{Kingma2014Adam} with default parameters ($lr=0.001$ and $\beta=(0.9,0.999)$).
	
	\subsection{Performance Comparison}
	\label{subsection:result_performance}
	
	Three datasets HPatches \cite{balntas2017hpatches}, Webcam \cite{verdie2015tilde}, Oxford \cite{mikolajczyk2005performance} are used to evaluate the performance. HPatches is divided into illumination changed subset and viewpoint changed subset, which are denoted as HP-i and HP-v respectively. Webcam contains sharp illumination changes but no viewpoint change. In this paper no method is trained on Webcam, so training set and testing set of Webcam are both available to evaluate performance, which is denoted as W-train and W-test. Oxford comprises of both illumination and viewpoint changes, but we don’t split it because the number of images in Oxford is small.
	
	Two used metrics Matching Score and Homography Estimation are identical to that used in \cite{detone2018superpoint} (see more details in Supplementary Section 4). Higher Matching Score and Homography Estimation are better. Briefly, Matching Score measures the ratio of the recovered ground truth correspondences over the number of points extracted by the detector in the shared viewpoint region. Homography Estimation measures the ability of an algorithm to estimate the homography that relates a pair of images by comparing it to ground truth homography. Similar to \cite{detone2018superpoint}, we fix correct distance $\epsilon=3$ pixels, and use RANSAC method implemented by Opencv Toolbox to estimate homography. In this paper all evaluations are performed on image size $640\times480$. To be fair we use the recommended hyperparameters for every method, except the maximum number of extracted interest points.  We keep no more than 1000 interest points for $640\times480$ images. 
	
	The performances of different methods are shown in Table \ref{tab:tabscore} and Table \ref{tab:tabhomo}.
	
	In summary, our Property Networks outperform the other methods on viewpoint changed subset of HPatches and Oxford, and achieve competitive performance on others according to Matching Score. With Homography Estimation, our methods outperform other methods on HPatches and Webcam and achieve competitive performance on Oxford. As the state-of-the-art algorithms, SIFT has low performance on illumination changed datasets and SuperPoint has no superiority on viewpoint changed datasets, but our Property Networks present much more stable results on all above datasets. 
	Furthermore, it’s interesting that SuperPoint has slight advantage on Webcam according to Matching Score but has large gap comparing to our PN-i-64 according to Homography Estimation. This is because Matching Score is a basic metric only reflecting the maximum number of points which are likely helpful to subsequent tasks. However, the space distribution of matched points is as important as the number of matched points for the final tasks such as homography estimation and camera pose estimation, which can’t be revealed by Matching Score. Sparsity is an essential property making the space distribution of interest point suit for these tasks. By improving sparsity and other properties jointly PN-i-64 achieve much higher performance for Homography Estimation on illumination changed dataset.
	
	Figure \ref{fig:figure2} demonstrates some visualization details about it. Whereas SuperPoint and SIFT achieve the best performance in some specific scenes, PN-i-64 model give reliable results under both illumination and viewpoint changes. Furthermore, above visualization results of PN-i-64 and SuperPoint explains the difference between Matching Score and Homography Estimation. 
	More results can be found in Supplementary Section 5.
	
	\noindent
	\begin{table}[h]
		\centering  
		\caption{Matching Score of Different Methods}
		\begin{tabular}{cccccc} 
			\toprule 
			&HP-i & HP-v& W-train& W-test& Oxford\\ 
			\midrule 
			SIFT& 0.295& 0.314& 0.130& 0.137& 0.357\\
			SURF& 0.300& 0.281& 0.128& 0.138& 0.395\\
			ORB& 0.335& 0.306& 0.141& 0.158& 0.421\\
			KAZE& 0.362& 0.276& 0.167& 0.182& 0.381\\
			LIFT& 0.336& 0.291& 0.172& 0.187& 0.325\\
			LF-Net& 0.299& 0.273& 0.159& 0.175& 0.326\\
			Super& \textbf{0.527}& 0.458& \textbf{0.317}& \textbf{0.330}& 0.434\\
			PN-i-64& 0.522& 0.472& 0.302& 0.316& 0.445\\
			PN-v-64& 0.466& \textbf{0.503}& 0.232& 0.249& \textbf{0.521}\\
			PN-128& 0.469& 0.464& 0.248& 0.259& 0.472\\
			\bottomrule 
			\label{tab:tabscore}
		\end{tabular} 
	\end{table}
	
	\subsection{Performance Analysis of Three PM Models}
	\label{subsection:result_variants}
	
	It’s an ultimate goal to learning an interest point model that can be generalized to all application scenes. From the viewpoint of actual application, it’s practical to learning an interest model to cope with specific conditions. The results of our three models confirm this opinion, which is intuitively demonstrated in Figure \ref{fig:figure3}.
	
	Though PN-128 has the highest potentials with largest description vector length, in our training it can’t converge well facing both sharp illumination changes and large range viewpoint changes, which makes it have no superiority for either illumination or viewpoint changes. From another perspective, if there are both sharp illumination changes and large range viewpoint changes in given application, PN-128 should be a reasonable tradeoff with limited learning ability. Note PN-128 achieves similar or better results comparing with SuperPoint whose length of description vector is 256.
	
	By focusing on sharp illumination changes and medium viewpoint changes, PN-i-64 achieves much better performance on illumination changed datasets, which demonstrates the flexibility of our training framework. What we need is only feeding the model corresponding images for different application scenes, without adjusting the architecture and objective anymore.
	PN-v-64 outperform PN-128 under viewpoint changes, but the superiority is relatively small. One reason is conventional convolution neural network such as Fully Convolutional Network can only achieve limited rotation invariant with convolution and pooling operation \cite{Cohen2016Group,Worrall2017Harmonic}. Explicitly estimating the orientation of interest point is a one of reliable ways to improve rotation invariant \cite{lowe2004distinctive,yi2016lift}.
	
	\noindent
	\begin{table}[h]
		\centering  
		\caption{Homography Estimation of Different Methods}
		\begin{tabular}{cccccc} 
			\toprule 
			&HP-i & HP-v& W-train& W-test& Oxford\\ 
			\midrule 
			SIFT& 0.842& 0.557& 0.526& 0.563& \textbf{0.650}\\
			SURF& 0.781& 0.444& 0.396& 0.42& 0.600\\
			ORB& 0.682& 0.334& 0.304& 0.329& 0.533\\
			KAZE& 0.809& 0.398& 0.449& 0.473& 0.492\\
			LIFT& 0.875& 0.462& 0.600& 0.628& 0.558\\
			LF-Net& 0.841& 0.435& 0.517& 0.573& 0.567\\
			Super& 0.938& 0.525& 0.713& 0.730& 0.592\\
			PN-i-64& \textbf{0.950}& 0.555& \textbf{0.802}& \textbf{0.811}& 0.608\\
			PN-v-64& 0.882& \textbf{0.572}& 0.564& 0.602& 0.642\\
			PN-128& 0.937& 0.554& 0.715& 0.730& 0.617\\
			\bottomrule 
			\label{tab:tabhomo}
		\end{tabular} 
	\end{table}

	\section{Conclusion}
	\label{section:conclusion}
	This paper proposes an entirely unsupervised training framework by maximizing the sparsity, repeatability and discriminability probability of interest point and its description. With Expectation Maximization algorithm and mini-batch approximation this framework can be optimized efficiently. As an implementation based on Fully Convolutional Network, Property Network outperforms state-of-the-art algorithms on a number of image matching dataset, which demonstrates the effectiveness and flexibility of our training framework. Future work will investigate more well-designed architecture of detector and descriptor which can reach more potential of this framework. Furthermore, our framework can be integrated with more properties to improve model performance in diverse applications, and some supervised information can also be formulated as properties which bring semantics to interest point.
	
	{\small
		\bibliographystyle{ieee}
		\bibliography{egbib}
	}

	\section*{Supplementary Material}
	In this supplementary material we give more details of our implementation and experimential results. In section \ref{section:MBEM} we implement the Expectation Maximization (EM) process with an efficient strategy called Mini-Batch approximation of EM (MBEM). Section \ref{section:arch} demonstrates the architecture of our Property Network. Section \ref{section:simu} outlines details of our simulation for illumination and viewpoint changes. In Section \ref{section:metric} we introduce performance metrics used in experiments and more experimential results are demonstrated in Section \ref{section:results}.
	
	\setcounter{section}{0}
	\section{Mini-Batch Approximation of Expectation Maxmization}
	\label{section:MBEM}
	\subsection{Problems of Original Expectation Maxmization}
	\label{subsection:MBEM_problem}
	Before discuss the difficultes of optimizing our objective with original Expectation Maxmization algorithm (EM), we first overview the objective of our training framework, which is formulated as
	\begin{equation}
	\left\{\begin{array}{c}\mathop{\arg\max}\limits_{\theta_{F}, \theta_{D}, y^1,.,y^K} P(\theta_{F}, \theta_{D},y^1,.,y^K)=\prod_{k} r\left(y^{k}\right) \cdot \hat{c}\left(y^{k}\right) \\ {s . t . N_{min}<\left\|y^{k}\right\|<N_{max}, s_{l o c}\left(y^{k}\right)=y^{k}}\end{array}\right.,
	\label{eq:latent_obj}
	\end{equation}
	where $\theta_F$ and $\theta_D$ are parameters of detector and descriptor respectively, and $y^1,...,y^K$ are \emph{latent variables}. Each $y^k$ is correspond to a scene $I^k$, and $y^k$ is a vector whose component $y_i^k$ is a binary random variable representing whether point $o_i^k$ satisfies desired properties. Point $o_i^k$ is defined as \emph{satisfied point} if $y_i^k=1$. Note $y^k$ need to be optimized because we don't known which points can become satisfied points. To be convenient for optimization we also define log-likelihood function
	\begin{equation}
	L=\log(P(\theta_{F}, \theta_{D},y^1,.,y^K)).
	\label{eq:expect_L_base}
	\end{equation}
	
	Because each $y^k$ is independent in objective (\ref{eq:latent_obj}) , we drop the superscript in the following parts of the supplement for the brief expression. Without ambiguity we always use $L$ to represent log-likelihood function no matter whether the $k$ is ignored or not.
	
	Properties considered in objective (\ref{eq:latent_obj}) are \emph{sparsity}, \emph{repeatability} and \emph{discriminability}. The repeatability and discriminability probability is
	\begin{equation}
	r(y) =\prod_i{\left(r_{i}\right)}^{y_{i}}\left( 1-r_{i} \right) ^{1-y_{i}},
	\label{eq:rep_redef}
	\end{equation}
	\begin{equation}
	\hat{c}(y)=\prod_{i}{(\hat{c}_i(y))^{y_i}},
	\label{eq:disc_redef}
	\end{equation}
	where $r_i$ and $\hat{c}_i(y)$ are probabilities that $o_i$ satisfies repeatability and discriminability respectively. The formulations of $r_i$ and $\hat{c}_i(y)$ can be found in main text. In this section we only need know $r_i$ don't rely on $y$, but $\hat{c}_i(y)$ depend on vector $y$.
	
	The constraint in objective (\ref{eq:latent_obj}) is named as \emph{sparsity constraint}.  $\left\|y\right\|=\sum_i{y_i}$ represents the number of satisfied points, and $[N_{min},N_{max}]$ are reasonable number range of interest points in a single scene. \emph{Local sparsity constraint} $s_{loc}$ is defined as
	\begin{equation}
	s_{loc}(y_i)=
	\begin{cases} 
	y_i,  & \left( \sum_{o_{i'} \in U(o_i)}y_{i'} \right)=0 \\
	0, & otherwise
	\end{cases},
	\label{eq:local_sparsity}
	\end{equation}
	where $U(o_i )$ represents the neighborhood of point $o_i$ whose radius is $rad$. Here $rad$ is a small integral (e.g., $rad=4$), and $U(o_i)$ doesn’t contain $o_i$ itself.
	
	Theoretically, objective (\ref{eq:latent_obj}) can be optimized with EM. Normally EM contains $T$ iterations, and an Expectation step (E-step) and a Maximization step (M-step) are conducted in each iteration. Let $\theta_F^0$ and $\theta_D^0$ are initial parameters, and the parameters of detector and descriptor are $\theta_F^t$ and $\theta_D^t$ after $t$ iterations. In each E-step we need obtain $p_i=P\left(y_i=1 | \theta_{F}^{t-1}, \theta_{D}^{t-1}\right)$ and compute the expectation of $L$ with respect to $y$ which is denoted as $E_y(L)$. In each M-step we need update $\theta_{F}^{t-1}$ and $\theta_{D}^{t-1}$ to $\theta_{F}^{t}$ and $\theta_{D}^{t}$ by maximizing $E_y(L)$.
	
	We first formulate $p_i$ theoretically. Without sparsity constraint the sample space of $y$ can be represented as $\{y^{(m)}\vert m=1,2,...,2^N\}$ where $N$ is the number of all the points in given scene. That's because the length of vector $y$ is $N$ and each component of $y$ is a binary variable. But with sparsity constraint this sample space should be more narrow. Denote the sample space of $y$ as $Y$ which can be formulated as
	\begin{equation}
	\begin{aligned}
	Y=
	&\left\{y^{(m)}\vert m=1,2,...,2^N,\right. \\
	&\left. N_{min}<\left\|y^{(m)}\right\|<N_{max}, s_{l o c}\left(y^{(m)}\right)=y^{(m)}\right\}.
	\end{aligned}
	\label{eq:possi_y_set}
	\end{equation}
	
	The sparsity constraint in objective (\ref{eq:latent_obj}) can be ignored if we ensure $y\in Y$. So when $y\in Y$ the distribution of $y$ can be reformulate as
	\begin{equation}
	\begin{aligned}
	P\left(y | \theta_{F}^{t-1}, \theta_{D}^{t-1}\right)
	&= r\left(y\right) \cdot \hat{c}\left(y\right),\\
	&=\prod_{i}{r_{i}^{y_{i}}\cdot(1-r_{i})^{1-y_{i}}\cdot\hat{c}_{i}(y)^{y_{i}}}
	\end{aligned}
	\label{eq:detail_prob_y}
	\end{equation} 
	Note we obtain (\ref{eq:detail_prob_y}) by substituting (\ref{eq:rep_redef}) and (\ref{eq:disc_redef}) into (\ref{eq:latent_obj}).
	
	Because the distribution $p_i$ is a marginal distribution of $P(y | \theta_{F}^{t-1}, \theta_{D}^{t-1})$, so we obtain
	\begin{equation}
	\begin{aligned}
	p_i
	&=P\left(y_i=1 | \theta_{F}^{t-1}, \theta_{D}^{t-1}\right)\\
	&=\frac{1}{Z}\sum_{y^{(m)} \in Y,y_i=1} P\left(y^{(m)} | \theta_{F}^{t-1}, \theta_{D}^{t-1}\right)\\
	&=\frac{1}{Z}\sum_{y^{(m)} \in Y_{y_i=1}} r(y^{(m)}) \cdot \hat{c}(y^{(m)}) \\
	&=\frac{1}{Z}\sum_{y^{(m)} \in Y_{y_i=1}}{\prod_{i}{r_{i}^{y_{i}^{(m)}}\cdot(1-r_{i})^{1-y_{i}^{(m)}}\cdot\hat{c}_{i}(y^{(m)})^{y_{i}^{(m)}}}}
	\label{eq:detail_solu_y}
	\end{aligned}
	\end{equation} 
	where $Z=\sum_{y^{(m)} \in Y} r(y^{(m)}) \cdot \hat{c}(y^{(m)})$ is the normalization factor, and $Y_{y_i=1}=\{y^{(m)}\vert y^{(m)}_i=1, y^{(m)}\in Y\}$. By combining \ref{eq:expect_L_base}, \ref{eq:rep_redef} and \ref{eq:disc_redef}, the expectation of $L$ with respect to $y$ is
	\begin{equation}
	\begin{aligned}
	E_{y}(L)=
	&\sum_{i} \bigl(p_i \log r_i+\left(1-p_i\right) \log \left(1-r_i\right)+\\
	&E_{y}(y_i\log(\hat{c}_i(y)))\bigr),
	\label{eq:expect_L}
	\end{aligned}
	\end{equation}
	Note we drop all the $k$ of $L$ and only considering single $y$. Here $\hat{c}_i(y)$ depend on $y$ so we can't straightforward simplify the expectation $E_{y}(y_i\log(\hat{c}_i(y)))$ in (\ref{eq:expect_L}).
	
	The $E_y(L)$ need be maximized in M-step, which can be achieved with Gradient Ascent algorithm (GA). According to definitions in main text, $r_i$ depends on $\theta_{F}$, and $\hat{c}_i(y)$ depends on $\theta_{D}$. The expressions of partial derivative are
	\begin{equation}
	\frac{\partial L}{\partial \theta^{t-1}_F} = \sum_i \frac{p_i-r_i}{r_i \cdot (1-r_i)} \cdot \frac{\partial r_i}{\partial \theta ^{t-1}_F},
	\label{eq:grad_F_1}
	\end{equation}
	\begin{equation}
	\frac{\partial L}{\partial \theta^{t-1}_D} = \sum_i \frac{\partial E_{y}(y_i\log(\hat{c}_i(y))}{\partial \theta_D^{t-1}}.
	\label{eq:grad_D_1}
	\end{equation}
	Here $\frac{\partial r_i}{\partial \theta_F}$ in (\ref{eq:grad_F_1}) can be computed directly, but the partial derivative in (\ref{eq:grad_D_1}) is more complex.
	
	Above E-step and M-step lead to high computational complexity from three aspects. 1) It’s inefficient to make GA achieve convergence on entire image set in each M-step (note (\ref{eq:grad_F_1}) and (\ref{eq:grad_D_1}) are defined for one scene but in fact we need compute them for all scenes). 2) Directly computing $p_i$ with (\ref{eq:detail_solu_y}) need traverse $y^{(m)}$ through entire $Y$, which leads to very high computational complexity. 3) Even with known $p_i$, computing (\ref{eq:grad_D_1}) still need traverse $y^{(m)}$ through entire $Y$.
	
	In the subsequent subsection, we solve above three problems with some approximations of original EM. The entire approximate algorithm is named as \emph{Mini-Batch approximation of Expectation Maxmization} (MBEM).
	
	\subsection{Considering Mini-Batch in Each Iteration}
	\label{subsection:MBEM_mini}
	Because it's too slow to make GA achieve convergence on entire image set in each M-step, in each iteration we only focus on a small subset of training set, which is normally named as \emph{mini-batch}.
	
	Here is more details. First we select size of mini-batch as $BS$ (in our training we fix $BS=2$). Construct $B_t=\{b \vert t\cdot BS+1\le b\le (t+1)BS\}$ in iteration $t$ and select mini-batch $\{I^k \vert k \in B_t\}$ to conduct E-step and M-step. Furthermore, each M-step only updates parameters once rather than multiple times until converge, because it’s not necessary to achieve convergence on a mini-batch which increases the risk of overfitting.
	
	\subsection{Efficient Approximation for Distribution of Latent Variable}
	\label{subsection:MBEM_appro_solu_y}
	Though distribution $p_i$ can be computed with (\ref{eq:detail_solu_y}), but this computation need traverse $y^{(m)}$ through entire sample space $Y$ which leads to very high computational complexity. There are two obstacles to simplify (\ref{eq:detail_solu_y}).
	\begin{enumerate}[1)]
		\item The definition of sample space $Y$ is integrated with sparsity constraint. It's easy to check whether a specific $y^{(m)}$ satisfies sparsity constraint or not, but it's difficult to straightforward obtain all satisfied $y^{(m)}$.
		\item To solve distribution $p_i$, we must consider all points because $\hat{c}_i$ depends on vector $y$.
	\end{enumerate}
	
	Whereas it's hard to solve (\ref{eq:detail_solu_y}) precisely, this subsection introduce an efficient strategy to approximate $p_i$ which comprises of three step. First, we approximate sample space $Y$ with $\hat{Y}$ which can be obtained efficiently. Second, the $\hat{c}_i(y)$ is approximated with $\hat{c}_i(\hat{y})$ where $\hat{y}$ is a constant for given $\theta_{F}$. Third, we simplify the approximate formulation of $p_i$.
	
	\subsubsection{Efficient Approximation of Sample Space}
	
	We first approximate $Y$ with $\hat{Y}$ to simplify the sparsity constraint. A reasonable $\hat{Y}$ should have two characteristics. First, all $y\in \hat{Y}$ satisfy sparsity constraint. That's mean $\hat{Y}$ can only be a subset of $Y$, i.e., we can only remove some elements from $Y$ in order to obtain $\hat{Y}$. Second, replace $Y$ with $\hat{Y}$ won't change $p_i$ significantly. That's mean elements in $\hat{Y}$ should have greater contributions to $p_i$ comparing to elements in $Y-\hat{Y}$.
	
	The first characteristic can be achieved by setting $y_{i'}=0$ for all $o_{i'}\in U(o_{i})$ if we want to setting $y_{i}=1$. The remaining question is how to select point $o_i$ to set $y_{i}=1$. According to (\ref{eq:detail_solu_y}), the larger $r_i$ and $\hat{c}_{i}(y)$ point $o_i$ has, the greater contribution $y_i=1$ can make. Considering $r_i$ can be obtained directly but $\hat{c}_{i}(y)$ depend on $y$, and we have assume $r_i$ and $\hat{c}_{i}(y)$ are independent in main text, so we can select point $o_i$ if $r_i$ is a local maximum and set $y_i=1$.
	According to above analyses we define
	\begin{equation}
	\hat{y}_i=
	\begin{cases}
	1, & r_i > max(r_{i'}\big|o_{i'}\in U(o_i))\\
	0, & otherwise
	\end{cases},
	\end{equation}
	where $max(\cdot)$ return the maximum of given set. Define $\hat{y}$ is a vector whose $i$th component is $\hat{y}_i$. Since the local maximum points that are deducted from the current model require the corresponding points in the sample $y^{m}$ must be also local maximum as $\hat{y}$, the sample space Y can be further reduced as
	\begin{equation}
	\hat{Y}=\left\{ y^{(m)} \big| N_{min}<\left\| y^{(m)} \right\|<N_{max},\;\; \forall i, y^{(m)}_i\le \hat{y}_i \right\} .
	\label{eq:appro_y_set}
	\end{equation}
	Note any $y \in \hat{Y}$ satifies $s_{loc} (y)=y$ becuase $s_{loc}(\hat{y})=\hat{y}$, this significantly benefit computation efficiency by avoid checking this constraint.
	
	\subsubsection{Efficient Approximation of $\hat{c}_i(y)$}
	
	The second problem is $\hat{c}_i$ depends on vector $y$ which make us have to consider all points when solve $p_i$. To reduce the computational complexity, we give an approximation of $\hat{c}_i(y)$ with $\hat{c}_i(\hat{y})$.
	We first make an assumption called
	
	\noindent
	\emph{Discriminability Consistency}:
	
	\emph{Given two feasible samples $y^{(1)},y^{(2)} \in \hat{Y}$ which satisfy $\{o_i \vert y^{(1)}_i=1 \}\subseteq  \{o_i \vert y^{(2)}_i=1 \}$, discriminability consistency 
		assumes $\forall i, i' \in \{o_i \vert y^{(1)}_i=1 \}$, if $\hat{c}_i(y^{(1)})<\hat{c}_{i'}(y^{(1)})$, then $\hat{c}_i(y^{(2)})<\hat{c}_{i'}(y^{(2)})$.}
	
	Discriminability consistency assumes if the discriminability of point $o_i$ is larger than $o_{i'}$ in satisfied point set $\{o_i \vert y_i=1 \}$, then inserting some interest points into this satisfied point set may not change their relative order. According to the definition $\hat{y}$ which is deduced from the local maximum $r$ computed with current model parameters, $\hat{y}$ must be the super set of all the other genuine interest point sets.
	
	If we view  any sample $y$ as $y^{(1)}$ and $\hat{y}$ as $y^{(2)}$, then the discriminability consistency will be satisfied. Our objective of optimization is to maintain the best descriminability of each interest point so that we can use $\hat{y}_i$ to replace $y_i$. Therefore,it's reasonable to select $\hat{y}$ to replace any sample  $y$ to guarantee discriminability consistency. Furthermore, the feasible samples are also replaced by $\hat{y}$ such that we only concern about the computation of $\hat{y}$. With $\hat{y}$ we can obtain 
	\begin{equation}
	\hat{c}(\hat{y})=\prod_{i} \exp \left(\alpha \hat{y}_i\left(h_i\left(\hat{y}\right)-H\right)\right),
	\label{eq:disc_base}
	\end{equation}
	where $h_i$ is the function computing disciminability for given point $o_i$, and its formulation can be found in main text.
	
	\subsubsection{Simplify the Formulation of $p_i$}
	
	First we give the formulation of $p_i$ combining above two approximations. Because $\hat{y}_i$ is a constant with given $r_i$, we denote $\hat{c}_i(\hat{y})$ as $\tilde{c}_i$ for the simplicity. Then (\ref{eq:detail_solu_y}) can be approximated with
	\begin{equation}
	p_i\approx\frac{1}{Z}\sum_{y^{(m)} \in \hat{Y}_{y_i=1}}{\prod_{i'}{r_{i'}^{y^{(m)}_{i'}}\cdot(1-r_{i'})^{1-y^{(m)}_{i'}}\cdot\tilde{c}_{i'}^{y^{(m)}_{i'}}}}.
	\label{eq:detail_appro_solo_y}
	\end{equation}
	Here $Z$ is still the normalization factor and we don't show its formulation repeatedly, and $Y_{y_i=1}$ in (\ref{eq:detail_solu_y}) is changed to $\hat{Y}_{y_i=1}=\{y^{(m)}\vert y_i=1, y^{(m)}\in \hat{Y}\}$.
	
	In order to ensure computation efficieny we need avoid the summation over $y^{(m)}$, the product over $i'$ and the computation of normalization factor $Z$ in (\ref{eq:detail_appro_solo_y}). In this subsection we intorduce our solution to achieve it.
	
	We make $z_i^{(m)}={r_i^{y_i^{(m)}}\cdot(1-r_i)^{1-y_i^{(m)}}\cdot\tilde{c}_i^{y_i^{(m)}}}$ for the simplicity , then (\ref{eq:detail_appro_solo_y}) can be reformulated as
	\begin{equation}
	\begin{aligned}
	p_i
	&=\frac{1}{Z}\sum_{y^{(m)} \in \hat{Y}_{y_i=1}}{ r_i\cdot \tilde{c}_i\prod_{i'\ne i}{z_{i'}^{(m)}}}\\
	&=\frac{1}{Z}r_i\cdot \tilde{c}_i\sum_{y^{(m)} \in \hat{Y}_{y_i=1}}{\prod_{i'\ne i}{z_{i'}^{(m)}}}.
	\end{aligned}
	\label{eq:reform_solo_y}
	\end{equation}
	Then we formulate $1-p_i$ according to (\ref{eq:detail_prob_y}), i.e.,
	\begin{equation}
	\begin{aligned}
	1-p_i
	&=P(y_i=0)\\
	&=\frac{1}{Z}\sum_{y^{(m)} \in \hat{Y}_{y_i=0}}{ (1-r_i)\prod_{i'\ne i}{z_{i'}^{(m)}}}\\
	&=\frac{1}{Z}(1-r_i)\sum_{y^{(m)} \in \hat{Y}_{y_i=0}}{\prod_{i'\ne i}{z_{i'}^{(m)}}},
	\end{aligned}
	\label{eq:reform_solo_no_y}
	\end{equation}
	where $\hat{Y}_{y_i=0}=\{y^{(m)}\vert y_i=0, y^{(m)}\in \hat{Y}\}$.
	
	Make $\|\cdot\|$ return the number of elements in given set, then (\ref{eq:reform_solo_y}) and (\ref{eq:reform_solo_no_y}) can be converted to
	\begin{equation}
	p_i=\frac{1}{Z}r_i\cdot \tilde{c}_i\cdot \left\|\hat{Y}_{y_i=1}\right\|\cdot avg_1\bigg(\prod_{i'\ne i}{z_{i'}}\bigg),
	\label{eq:solo_y_avg}
	\end{equation}
	\begin{equation}
	1-p_i=\frac{1}{Z}(1-r_i)\cdot \left\|\hat{Y}_{y_i=0}\right\|\cdot avg_0\bigg(\prod_{i'\ne i}{z_{i'}}\bigg).
	\label{eq:solo_no_y_avg}
	\end{equation}
	where 
	\begin{equation}
	avg_1\bigg(\prod_{i'\ne i}{z_{i'}}\bigg)=\frac{1}{\left\|\hat{Y}_{y_i=1}\right\|}\sum_{y^{(m)} \in \hat{Y}_{y_i=1}}{\prod_{i'\ne i}{z_{i'}^{(m)}}},
	\end{equation}
	\begin{equation}
	avg_0\bigg(\prod_{i'\ne i}{z_{i'}}\bigg)=\frac{1}{\left\|\hat{Y}_{y_i=0}\right\|}\sum_{y^{(m)} \in \hat{Y}_{y_i=0}}{\prod_{i'\ne i}{z_{i'}^{(m)}}}.
	\end{equation}
	
	Whether $y_i=1$ or $y_i=0$ only slightly change $\prod_{i'\ne i}{z_{i'}}$, and this difference can be ignored after average them over $\hat{Y}_{y_i=1}$ and $\hat{Y}_{y_i=0}$, so $avg\left(\hat{Y}_{y_i=1}\right) \approx avg\left(\hat{Y}_{y_i=0}\right)$. With (\ref{eq:solo_y_avg}) and (\ref{eq:solo_no_y_avg}) we can obtain
	\begin{equation}
	\begin{aligned}
	p_i
	&=\frac{p_i}{p_i+(1-p_i)}\\
	&\approx\frac{r_i\cdot \tilde{c}_i\cdot \left\|\hat{Y}_{y_i=1}\right\|}{r_i\cdot \tilde{c}_i\cdot \left\|\hat{Y}_{y_i=1}\right\|+(1-r_i)\cdot \left\|\hat{Y}_{y_i=0}\right\|}
	\label{eq:solo_y_effic}
	\end{aligned}
	\end{equation}
	Note $r_i$ and $\tilde{c}_i$ can be computed efficiently without interaction to $\hat{Y}$. So only $\|\hat{Y}_{y_i=1}\|$ and $\|\hat{Y}_{y_i=0}\|$ need to be determined in (\ref{eq:solo_y_effic}).
	
	We first compute $\|\hat{Y}\|$ with (\ref{eq:appro_y_set}). Each $y$ in $\hat{Y}$ corresponds to a set of satisfied points which is select from $\{ o_i\vert \hat{y}_i=1\}$. This selection can be splited into two steps. First, we determine the number of satisfied points, i.e., select $n \in [N_{min},N_{max}]$. Second, we choose $n$ point from $\{ o_i\vert \hat{y}_i=1\}$, and the corresponding $y$ is obtained. According to above two step $\|\hat{Y}\|$ can be calculated with
	\begin{equation}
	\begin{aligned}
	\left\|\hat{Y}\right\|
	&=C_{\|\hat{y}\|}^n\\
	&=\sum_{n=N_{min}}^{N_{max}}\frac{\|\hat{y}\|!}{n!(\|\hat{y}\|-n)!},
	\label{eq:total_y_num}
	\end{aligned}
	\end{equation}
	where $C_{\|\hat{y}\|}^n$ indicate the number of combinations.
	
	In computation of $\|\hat{Y}_{y_i=1}\|$, we can only choose $\|y\|-1$ satisfied points because we have select $o_i$ as a satisfied point. So
	\begin{equation}
	\begin{aligned}
	\left\|\hat{Y}_{y_i=1}\right\|
	&=C_{\|\hat{y}\|-1}^{n-1}\\
	&=\sum_{n=N_{min}-1}^{N_{max}-1}\frac{(\|\hat{y}\|-1)!}{n!(\|\hat{y}\|-n)!}.
	\label{eq:y1_num}
	\end{aligned}
	\end{equation}
	Then 
	\begin{equation}
	\left\|\hat{Y}_{y_i=0}\right\|=\left\|\hat{Y}\right\|-\left\|\hat{Y}_{y_i=1}\right\|.
	\label{eq:y0_num}
	\end{equation}
	
	Substituting (\ref{eq:y1_num}), (\ref{eq:y0_num}) and definitions of $r_i$ and $c_i$ (details can be found in main text) into (\ref{eq:solo_y_effic}), the approximate $p_i$ can be computed efficiently.
	
	\subsection{Efficient Approximation for Partial Derivative of Description Parameters}
	\label{subsection:MBEM_derivative}
	In Subsection \ref{subsection:MBEM_appro_solu_y} we can obtain distribution $p_i$ efficiently.
	But even with known $p_i$ computing $E_{y}(y_i\log(\hat{c}_i(y))$ and its partial derivative in (\ref{eq:grad_D_1}) still need traverse $y$ through entire $Y$ because $\hat{c}_i$ depends on $y$. Fortunately, in Subsection \ref{subsection:MBEM_appro_solu_y} we have solve this problem by approximating $y$ with $\hat{y}$ under the assumption of discriminability consistency. Because $\hat{y}$ is independent of $y$, so
	\begin{equation}
	\begin{aligned}
	E_{y}(y_i\log(\hat{c}_i(\hat{y}))
	&=\log(\hat{c}_i(\hat{y}))E_{y}(y_i)\\
	&=p_i\log(\tilde{c}_i),
	\end{aligned}
	\label{eq:grad_E_effic}
	\end{equation}
	where $\tilde{c}_i=\hat{c}_i(\hat{y})$ is defined in Subsection \ref{subsection:MBEM_appro_solu_y}. Substituting (\ref{eq:grad_E_effic}) into (\ref{eq:grad_D_1}) we obtain
	\begin{equation}
	\begin{aligned}
	\frac{\partial L}{\partial \theta^{t-1}_D} 
	&= \sum_i \frac{\partial E_{y}(y_i\log(\hat{c}_i(y))}{\partial \theta_D^{t-1}}\\
	&\approx \sum_i p_i\frac{\partial \log(\tilde{c}_i)}{\partial \theta_D^{t-1}}\\
	&= \sum_i \frac{p_i}{\tilde{c}_i} \frac{\partial \tilde{c}_i}{\partial \theta_D^{t-1}}.
	\label{eq:grad_D_effic}
	\end{aligned}
	\end{equation}
	According to the definition of discriminability in main text, $\frac{\partial \tilde{c}_i}{\partial \theta_D^{t-1}}$ can be computed directly. So the entire Gradient Ascent algorithm can be performed.
	
	\section{Architecture of Property Network}
	\label{section:arch}
	
	\begin{figure}
		\begin{minipage}[b]{0.47\textwidth}
			\centering
			\includegraphics[width=1\textwidth]{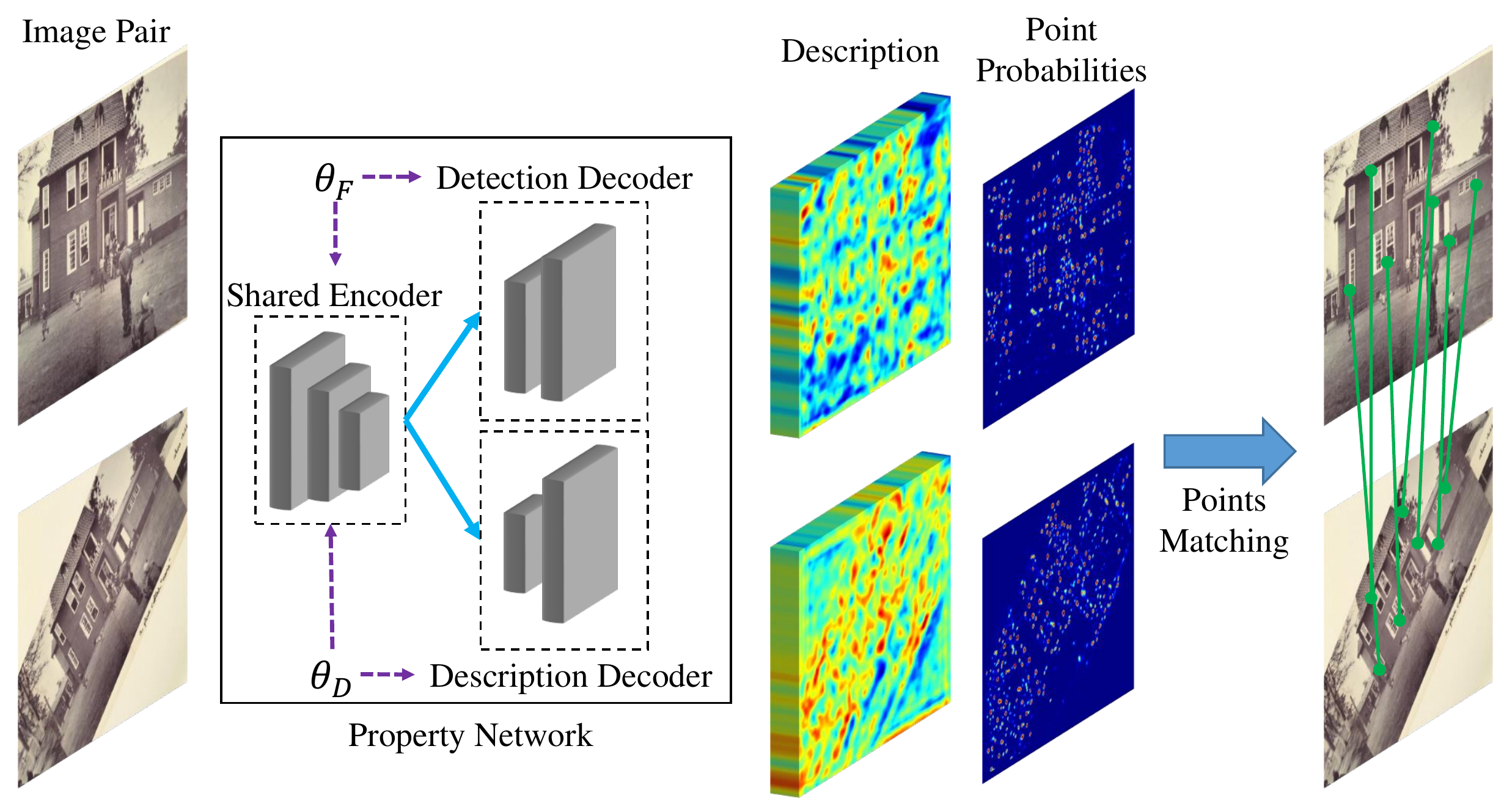}
			\caption{Overview of Property Network.}
			\label{fig:figure1}
		\end{minipage}
		
		\begin{minipage}[b]{0.47\textwidth}
			\centering
			\includegraphics[width=1\textwidth]{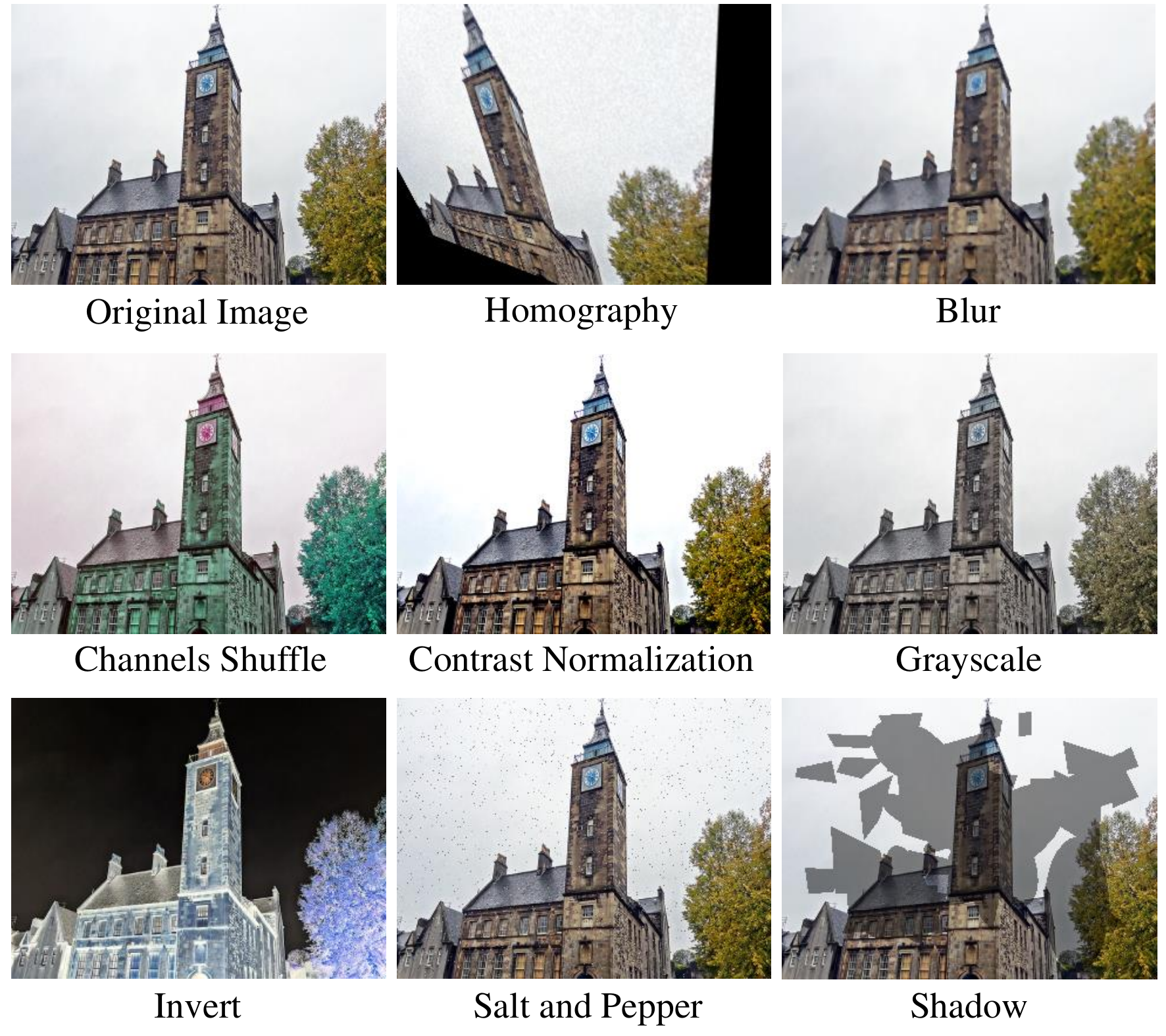}
			\caption{Examples of simulations for illumination and viewpoint changes.}
			\label{fig:figure2}
		\end{minipage}
	\end{figure}
	Property Network (PN) is a specific implementation of our training framework. In PN, both detector and descriptor are implemented with Fully Convolutional Network \cite{long2015fully,zeiler2010deconvolutional}, whose architecture is inspired by \cite{detone2018superpoint}. Figure \ref{fig:figure1} shows the architecture of PN briefly. Detector comprises of encoder and detection decoder, and descriptor comprises of encoder and description decoder. Both detector and descriptor share the same encoder. Table \ref{tab:arch} denomstrate more details of architecture of PN.
	
	In Table \ref{tab:arch}, We ignore all Batch Normalization and ReLU layers, which follow every convolution layer except the final convolution layer in Conv5 and Conv4-2. “Concatenate” operation concatenates current feature map and given feature map into a new feature map. "trans-conv" indicate transposed convolution \cite{zeiler2010deconvolutional}. p=1 means padding is 1, and s=1 indicates the strider of sliding window is 1. $des\_len$ is the length of description vector. As introduced in main text, we make $des\_len=64$ in PN-i-64 and PN-v-64, and $des\_len=128$ in PN-128.
	
	\noindent
	\begin{table}[h]
		\centering
		\caption{Architecture of Property Network}
		\begin{tabular}{ccc}
			\toprule
			Part& Block& Layer Details\\
			\midrule
			\multirow{8}*{encoder} & \multirow{2}*{Conv1}& conv, $3\times 3\times 32$, p=1, s=1 \\
			~ & ~ &  conv, $3\times 3\times 32$, p=1, s=1 \\
			~ &  & max-pooling, $2\times 2$, strider 2 \\
			~ & \multirow{2}*{Conv2}& conv, $3\times 3\times 64$, p=1, s=1 \\
			~ & ~ &  conv, $3\times 3\times 64$, p=1, s=1 \\
			~ &  & max-pooling, $2\times 2$, strider 2 \\
			~ & \multirow{2}*{Conv3}& conv, $3\times 3\times 64$, p=1, s=1 \\
			~ & ~ &  conv, $3\times 3\times 64$, p=1, s=1 \\
			\midrule
			\multirow{9}*{detection} & \multirow{3}*{Conv4-1}& trans-conv, $3\times 3\times 64$, p=1, s=2 \\
			~ & ~ &  concatenate with output of Conv2 \\
			~ &  & conv, $3\times 3 \times 64$, p=1, s=1 \\
			~ & \multirow{4}*{Conv5}& trans-conv, $3\times 3\times 64$, p=1, s=2 \\
			~ & ~ &  concatenate with output of Conv1 \\
			~ & ~ & conv, $3\times 3\times 32$, p=1, s=1 \\
			~ & ~ & conv, $3\times 3\times 1$, p=1, s=1 \\
			~ & & sigmoid function \\
			~ & NMS & conv, max-pooling, $7\times 7$, strider 1 \\
			\midrule
			\multirow{5}*{description} & \multirow{3}*{Conv4-2}& conv, $3\times 3\times 64$, p=1, s=1 \\
			~ & ~ &  conv, $3\times 3\times 64$, p=1, s=1 \\
			~ &  & conv, $3\times 3\times des\_len$, p=1, s=1 \\
			~ & Norm2 & $L2$ Normalization for each point \\
			~ & upsample & upsample with factor=4 \\
			\bottomrule
			\label{tab:arch}
		\end{tabular}
	\end{table}
	
	\section{Simulations for Viewpoint and Illumination Changes}
	\label{section:simu}
	In this section we outline our simulations for illumination and viewpoint changes. For illumination simulation we randomly select the transformations to change pixel value. For viewpoint simulation we randomly generate homography matrices used to perform homography transformation.  Figure \ref{fig:figure2} shows several examples of our simulated results.
	We first introduce more details of simulations for illunimation changes. Our slmulations comprise of seven kinds of transformations. In the discussion below we assume the original image have three color channels with 256 gray levels.
	\begin{enumerate}[1)]
		\item Image Blur. Apply Gaussian blur, average blur or median bluron image.
		\item Channels Shuffle. Permute the order of the color channels of image.
		\item Contrast Normalization. Change the contrast in images by moving pixel values away or closer to 128.
		\item Grayscale. Convert images to grayscale and mixe with the original image with a random weight.
		\item Invert all pixels in given image, i.e. set them to $255-original\_pixel\_value$.
		\item Salt and Pepper noise. Randomly replace some pixels with very white or black colors.
		\item Shadow. Randomly insert some dark shapes into image.
	\end{enumerate}
	
	In main text we have mentioned different PN models are trained with different level of simulations. PN-v-64 is trained with only Image Blur, Contrast Normalization and Shadow, and PN-i-64 and PN-128 are trained with all kind of simulations for illumination changes.
	
	In simulations for viewpoint changes, we randomly generate homography matrices and use them to perform homography transformation. Different PN models are also trained with different level of homography transformation. In training of PN-i-64, we ensure the rotation angle of homography transformation to be less than $45^{\circ}$, and for PN-v-64 and PN-128 we don't conduct restriction of rotation angle.
	
	All above simulations are online, i.e., in each training iteration we randomly perform above transformations on current image mini-bacth, and all transformed images will be discarded after this iteration.
	
	\section{Performance Metrics}
	\label{section:metric}
	Two metrics used in our experiment are Matching Score and Homography Estimation, which are identical to that used in \cite{detone2018superpoint}. In this section we introduce their definitions.
	\subsection{Matching Score}
	Matching Score measures the overall performance of  interest point detector and descriptor. It measures the ratio of ground truth correspondences that can be recovered by detector and descriptor over the number of interest point proposed by the pipeline in the shared viewpoint
	region. Suppose the detector extracts $\| IP_1\|$ points from image $I_1$ and $\| IP_2\|$ points from image $I_2$ in the shared viewpoint
	region of $I_1$ and $I_2$. With descriptions of extracted points and a specific matching strategy, suppose $corr_{12}$ points in $I_1$ are correctly matched to points in $I_2$, and $corr_{21}$ points in $I_2$ are correctly matched to points in $I_1$. Then Matching Score for this pair of image is
	\begin{equation}
	M\text{-}score=\frac{1}{2}\left(\frac{corr_{12}}{\| IP_1\|}+\frac{corr_{21}}{\| IP_2\|}\right).
	\end{equation}
	Same to \cite{detone2018superpoint}, we use two-way nearest neighbor matching as the matching strategy. And Matching Score for entire dataset is obtained by averaging $M\text{-}score$ over all image pairs.
	
	\subsection{Homography Estimation}
	Homography Estimation measures the ability of an algorithm to estimate the homography that relates a pair of images by comparing it to ground truth homography. Denote the estimated homography as $\hat{H}$ and ground truth homography as $H$. Be identical to \cite{detone2018superpoint}, Homography Estimation is defined with the error of four corners of one image onto the other. Denote the four corners of
	the first image as $c_1$, $c_2$, $c_3$ and $c_4$. We then apply the ground truth $H$ to get the ground truth corners in the second image which is denote as $c_1'$, $c_2'$, $c_3'$ and $c_4'$, and the estimated homography $\hat{H}$ to get the estimated corners in the second image which is denoted as $\hat{c_1}'$, $\hat{c_2}'$, $\hat{c_3}'$ and $\hat{c_4}'$. And the error of homography is defined as
	\begin{equation}
	Homo\_error=\frac{1}{4}\sum_{i=1}^4\| \hat{c_i}'-c_i'\|.
	\end{equation}
	And the measure of Homography Estimation is defined as
	\begin{equation}
	HE=\mathbb{I}(Homo\_error\le \epsilon).
	\end{equation}
	Here $\mathbb{I}$ is the indicator function, and $\epsilon$ is the threshold to judge the correctness of homography. Same to \cite{detone2018superpoint} we set $\epsilon=3$ in our experiment.
	
	\section{More Experimential Results}
	\label{section:results}
	\begin{figure*}
		\begin{center}
			\includegraphics[width=1.0\textwidth]{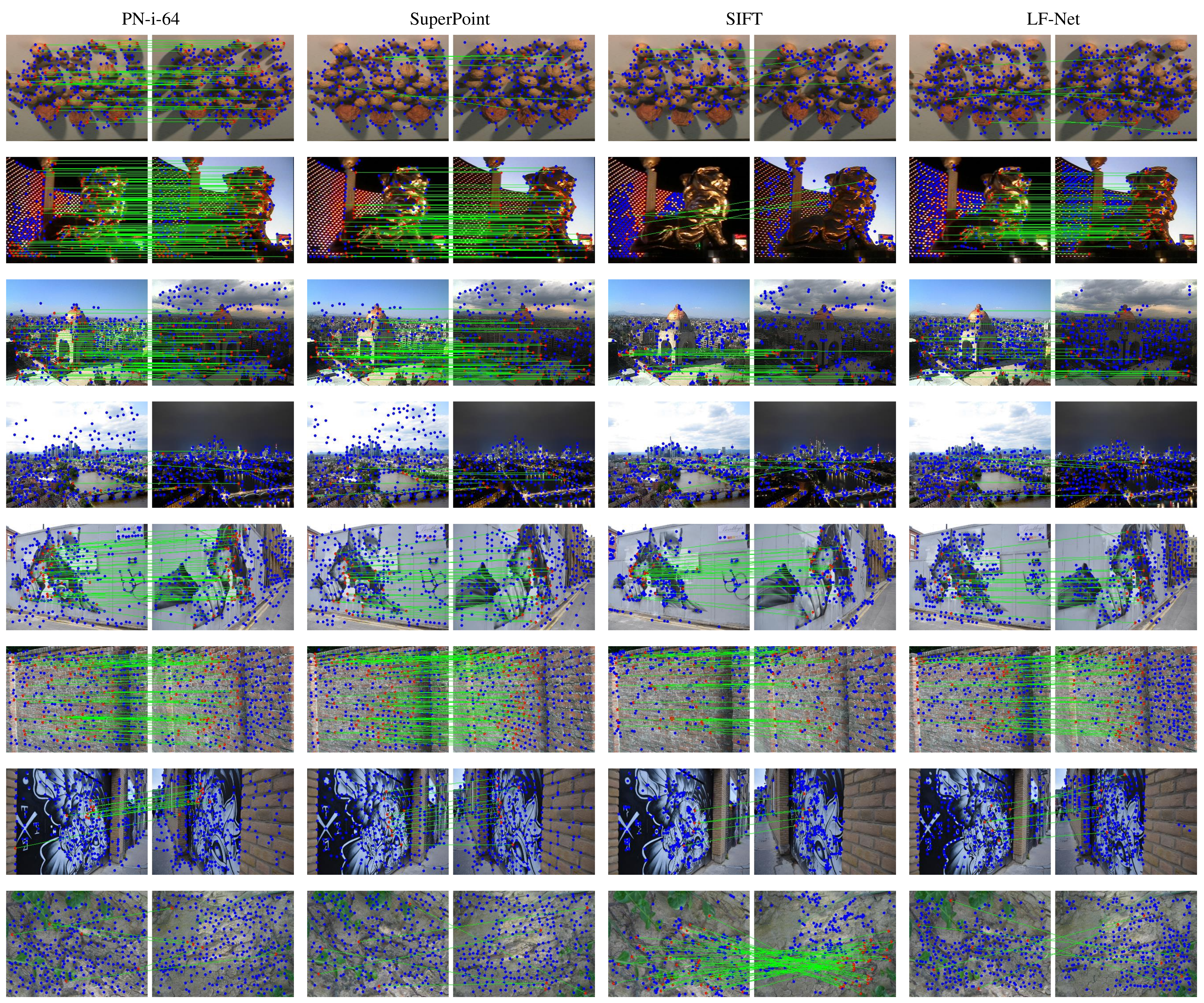}
		\end{center}
		\caption{Visual matching results of state-of-the-art algorithms and our PN-i-64 model (M-score indicates Matching Score, and Homo-error means the error of estimated Homography) . First col: our PN-i-64 model, second col: SuperPoint, third col: SIFT and fourth col: LF-Net. All points extracted by each method are shown in blue dots whereas correct matches are shown in green lines and red dots.}
		\label{fig:figure3}
	\end{figure*}
	
	In main text we have demostrated results on image size $640\times 480$. In this subsection we give results of different methods on image size $320\times 240$. All experiment configures are same as that for image size $640\times 480$ except the maximum number of extracted interest points.  We keep no more than 300 interest points for $320\times240$ images. Table \ref{tab:tabscore} and \ref{tab:tabhomo} demostrate Matching Score and Homography Estimation of different results. And Figure \ref{fig:figure3}  shows some matching results with image size $320\times 240$. Here our PN-i-64, PN-v-64 and PN-128 are identical to what in main text.

	\noindent
	\begin{table}[h]
		\centering
		\caption{Matching Score of Different Methods on Image Size $320\times240$}
		\begin{tabular}{cccccc}
			\toprule
			&HP-i & HP-v& W-train& W-test& Oxford\\
			\midrule
			SIFT& 0.311& 0.299& 0.141& 0.151& 0.405\\
			SURF& 0.331& 0.289& 0.147& 0.162& 0.44\\
			ORB& 0.361& 0.284& 0.165& 0.185& 0.447\\
			KAZE& 0.398& 0.288& 0.189& 0.205& 0.422\\
			LIFT& 0.374& 0.291& 0.202& 0.213& 0.387\\
			LF-Net& 0.372& 0.31& 0.206& 0.219& 0.429\\
			Super& \textbf{0.586}& 0.462& \textbf{0.358}& \textbf{0.374}& 0.496\\
			PN-i-64& 0.565& 0.466& 0.336& 0.351& 0.483\\
			PN-v-64& 0.512& \textbf{0.513}& 0.261& 0.278& \textbf{0.572}\\
			PN-128& 0.516& 0.456& 0.274& 0.288& 0.516\\
			\bottomrule
			\label{tab:tabscore}
		\end{tabular}
	\end{table}
	
	\noindent
	\begin{table}[h]
		\centering
		\caption{Homography Estimation of Different Methods on Image Size $320\times240$}
		\begin{tabular}{cccccc}
			\toprule
			&HP-i & HP-v& W-train& W-test& Oxford\\
			\midrule
			SIFT& 0.791& 0.585& 0.453& 0.461& \textbf{0.8}\\
			SURF& 0.716& 0.353& 0.314& 0.332& 0.675\\
			ORB& 0.611& 0.185& 0.237& 0.228& 0.475\\
			KAZE& 0.725& 0.336& 0.334& 0.362& 0.592\\
			LIFT& 0.826& 0.395& 0.484& 0.517& 0.658\\
			LF-Net& 0.802& 0.384& 0.49& 0.531& 0.675\\
			Super& 0.911& 0.493& 0.664& 0.684& 0.642\\
			PN-i-64& \textbf{0.958}& 0.572& \textbf{0.764}& \textbf{0.812}& 0.717\\
			PN-v-64& 0.864& 0.584& 0.495& 0.556& 0.792\\
			PN-128& 0.918& \textbf{0.588}& 0.654& 0.697& 0.775\\
			\bottomrule
			\label{tab:tabhomo}
		\end{tabular}
	\end{table}
	
\end{document}